\documentclass[11pt]{article}
\usepackage[utf8]{inputenc}
\usepackage[T1]{fontenc}
\usepackage{amsmath,amssymb}
\usepackage{graphicx}
\usepackage{tabularx}
\usepackage{multirow}
\usepackage{booktabs}
\usepackage{hyperref}
\usepackage{xcolor}
\usepackage{caption}
\usepackage{subcaption}
\usepackage{tikz}
\usepackage{pgfplots}
\pgfplotsset{compat=1.18}
\usetikzlibrary{positioning,arrows,shapes,fit}
\usepackage{placeins}
\usepackage{fancyhdr}
\usepackage[numbers]{natbib}   
\usepackage{array} 
\usepackage{makecell} 
\DeclareMathOperator*{\argmax}{arg\,max}

\title{Creative Adversarial Testing (CAT): A Novel Framework for Evaluating Goal-Oriented Agentic AI Systems}
\author{Hassen Dhrif \\ Amazon, Bellevue WA, USA \\ \texttt{hdhrif@amazon.com}}

\date{}

\begin{document}
\maketitle

\begin{abstract}
Agentic AI represents a paradigm shift in enhancing the capabilities of generative AI models. While these systems demonstrate immense potential and power, current evaluation techniques primarily focus on assessing their efficacy in identifying appropriate agents, tools, and parameters. However, a critical gap exists in evaluating the alignment between an Agentic AI system's tasks and its overarching goals. This paper introduces the Creative Adversarial Testing (CAT) framework, a novel approach designed to capture and analyze the complex relationship between Agentic AI tasks and the system's intended objectives.

We validate the CAT framework through extensive simulation using synthetic interaction data modeled after Alexa+ audio services, a sophisticated Agentic AI system that shapes the user experience for millions of users globally. This synthetic data approach enables comprehensive testing of edge cases and failure modes while protecting user privacy. Our results demonstrate that the CAT framework provides unprecedented insights into goal-task alignment, enabling more effective optimization and development of Agentic AI systems.

\end{abstract}
\maketitle

\section{Introduction}

Agentic AI refers to systems—often based on large language models (LLMs)—that perceive their environment and act autonomously to achieve goals, going beyond passive text generation. In formal terms, an autonomous agent is “a system situated within and a part of an environment that senses that environment and acts on it, over time, in pursuit of its own agenda”~\cite{wang2025survey}. Recent advances show that scaled-up LLMs can approximate such agents: for example, \textit{Tree of Thoughts}~\cite{yao2023tree} extends chain-of-thought prompting to allow LLMs to explore multiple reasoning paths and deliberately plan, dramatically improving problem-solving on complex tasks. Likewise, \textit{Reflexion}~\cite{shinn2023reflexion} equips an LLM agent with verbal self-reflection: the agent generates textual feedback from trial results and reuses it to refine future actions, yielding large gains (e.g., boosting HumanEval pass@1 from 80\% to 91\%).

Unlike purely generative models, agentic LLMs can iteratively plan and adapt. For instance, \textit{AdaPlanner}~\cite{sun2023adaplanner} uses a closed-loop framework in which an LLM agent generates an initial plan and then adaptively refines it using environmental feedback. \textit{SWIFTSAGE}~\cite{lin2023swiftsage} adopts a dual-process architecture: a fast ``Swift'' LM proposes actions by behavior cloning, while a slow ``Sage'' LLM (e.g., GPT-4) conducts deliberate subgoal planning. This combination of intuition and deliberation significantly outperforms single-strategy baselines on complex interactive reasoning tasks. Other recent examples include \textit{DEPS}~\cite{wang2023deps}, which interleaves planning with self-explanation to improve multi-task agents, and \textit{Prospector}~\cite{kim2024prospector}, which pairs an LLM ``Actor'' with a fine-tuned ``Critic'' to rank trajectories, both boosting agentic performance in benchmarks.

To evaluate such agentic systems, new benchmarks and datasets have been developed. \textit{Mind2Web}~\cite{deng2023mind2web} provides over 2{,}000 tasks across 137 real websites, testing a web-crawling agent’s ability to follow instructions on arbitrary sites. \textit{InterCode}~\cite{yang2023intercode} frames code generation as a sequential decision process with execution feedback, enabling LLMs to interactively debug and refine code. Research on multi-agent coordination has also been explored: for example, \textit{MultiPrompter}~\cite{kim2023multiprompter} trains cooperative LLM ``prompters'' via multi-agent reinforcement learning to collaboratively construct prompts, yielding higher-quality outputs (e.g., in text-to-image tasks) than independent prompting.

Together, these developments illustrate the rise of agentic AI: LLM-based agents that plan, act, and learn from feedback in complex environments. In this paper, we update the foundational discussion by incorporating recent peer-reviewed work (2023--2024) in this space. We survey definitions and early theories of agency, and then review modern architectures and benchmarks where LLMs act as autonomous agents. This sets the stage for our contributions in advancing agentic AI.

\section{Related Work}

Recent research on LLM-based agents can be grouped into several threads:

\textbf{Deliberative Planning with LLMs:}
Building on chain-of-thought prompting~\cite{wei2022cot}, methods like \textit{Tree of Thoughts}~\cite{yao2023tree} enable search over intermediate ``thoughts'' so that the model considers multiple future paths. \textit{DEPS}~\cite{wang2023deps} alternates between planning and self-explanation steps, using failure feedback to iteratively refine plans in open-world tasks. These approaches aim to overcome the myopic nature of one-shot prompting by adding structured reasoning.

\textbf{Learning from Feedback:}
Several works equip LLM agents with feedback loops akin to reinforcement learning. \textit{Reflexion}~\cite{shinn2023reflexion} uses natural-language feedback: after each trial, the agent generates a textual summary of success or failure, which it then uses to guide future choices. \textit{AdaPlanner}~\cite{sun2023adaplanner} takes environment observations as input, adaptively updating its plan step-by-step. \textit{Prospector}~\cite{kim2024prospector} further formalizes this idea by using an LLM ``Critic'' (trained on reward signals) to rank the actor’s generated action sequences, combining in-context learning with fine-tuning. These methods demonstrate that LLM agents can improve over time without updating model weights directly.

\textbf{Agent Architectures for Interactive Tasks:}
Other works propose specialized agent frameworks. \textit{SWIFTSAGE}~\cite{lin2023swiftsage} explicitly integrates two systems of thought (fast LM vs. slow LM) to tackle tasks requiring hierarchical planning. \textit{Mind2Web}~\cite{deng2023mind2web} and \textit{InterCode}~\cite{yang2023intercode} focus on environment-specific challenges: the former builds a generalist web navigation agent, while the latter creates an interactive coding benchmark where the agent writes and executes code in a sandbox. These benchmarks highlight different aspects of agency, from real-world perception to multimodal decision-making.

\textbf{Collaborative and Modular Agents:}
Moving beyond single-agent setups, some work explores multi-agent coordination. \textit{MultiPrompter}~\cite{kim2023multiprompter} divides a large prompt optimization problem among cooperative ``prompter'' agents using multi-agent RL, improving performance in multimodal tasks. More generally, the use of multiple specialized LMs (e.g., actor/critic, fast/slow) points toward modular agent designs.

\textbf{Surveys and Definitions:}
Several recent surveys contextualize these trends. Wang et al.~\cite{wang2025survey} define autonomous agents in the context of LLMs and categorize emerging LLM-based agents from a holistic perspective. Unlike older RL-based agent work, LLM agents draw on massive pretrained knowledge and emergent reasoning abilities, enabling new forms of human-like decision making.

Overall, the landscape of agentic AI is rapidly evolving. We build on these developments by ensuring our references reflect the latest peer-reviewed contributions (especially from top venues in 2023--2024) and by situating our work within this updated context.

\section{Methodology}
The Creative Adversarial Testing (CAT) framework introduces a novel approach to AI system evaluation through a hierarchical architecture that systematically bridges the gap between task execution and goal achievement. This section presents the framework's theoretical foundations, core mechanisms, and their practical application in voice-activated audio services.

\subsection{Framework Architecture}
\begin{figure*}[t]
    \centering
    \begin{minipage}{0.31\textwidth}
        \centering
        \includegraphics[width=0.6\textwidth]{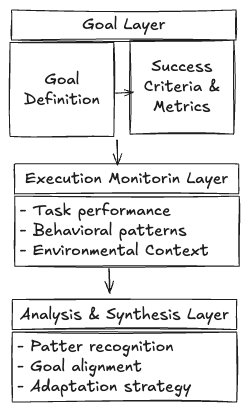}
        \caption{CAT Framework}
        \label{fig:1_framework_goal}
    \end{minipage}%
    \hfill
    \begin{minipage}{0.31\textwidth}
        \centering
        \includegraphics[width=0.8\textwidth]{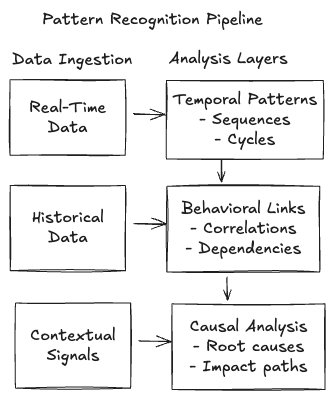}
        \caption{Multi-dimensional Pattern Analysis System}
        \label{fig:2_pattern_evolution}
    \end{minipage}%
    \hfill
    \begin{minipage}{0.31\textwidth}
        \centering
        \includegraphics[width=0.7\textwidth]{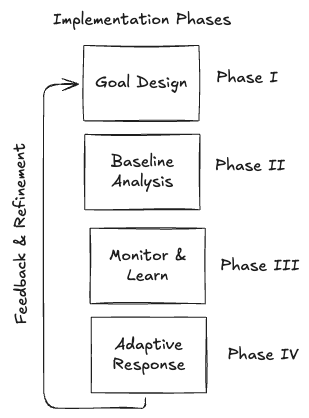}
        \caption{Implementation Framework}
        \label{fig:3_implementation}
    \end{minipage}
\end{figure*}

The CAT framework employs a three-layer architecture designed to transform granular task-level metrics into meaningful goal-oriented outcomes, as illustrated in Figure~\ref{fig:1_framework_goal}. Each layer serves a specific purpose in the evaluation pipeline:

1) The Goal Layer establishes the evaluation framework through explicit goal formulation and success criteria definition. In the context of audio services, high-level goals might include "enhance music discovery experience" or "increase podcast completion rates." These goals are structured hierarchically:
   - Strategic: "Build sustainable user engagement patterns"
   - Tactical: "Improve content discovery accuracy"
   - Operational: "Reduce irrelevant recommendations"

2) The Execution Monitoring Layer continuously observes system behavior across multiple dimensions. As shown in Figure~\ref{fig:2_pattern_evolution}, this layer implements sophisticated pattern recognition algorithms to identify relationships between individual actions (e.g., voice commands, content selections) and their contributions to goal achievement (e.g., sustained listening sessions, content completion).

3) The Integration Layer combines insights from multiple evaluation streams into actionable metrics.

\subsection{Goal-Task Alignment Mechanism}
The core challenge in evaluating AI systems lies in quantifying how well task-level performance translates to meaningful goal achievement. Consider a voice-activated music discovery scenario:
- Task: Accurate recognition of the command "play something similar"
- Goal: Help users discover new music they genuinely enjoy

To address this alignment challenge, we introduce the Goal Achievement Index (GAI):

\begin{equation}
    \text{GAI} = \frac{\sum(w_i \cdot T_i \cdot G_i)}{\sum(w_i \cdot T_i)}, \quad T_i, G_i \in [0,1], \sum w_i = 1
    \label{eq:gai}
\end{equation}

where:
- $T_i$ represents normalized task performance metrics (e.g., command recognition accuracy: 0.95)
- $G_i$ quantifies goal progress indicators (e.g., music discovery satisfaction: 0.73)
- $w_i$ are dynamic weights that adjust based on context (e.g., peak listening hours: 0.8)

For example, in podcast discovery:
\begin{table}[h]
    \centering
    \caption{Goal-Task Alignment Example}
    \begin{tabular}{lll}
        \toprule
        Task ($T_i$) & Goal ($G_i$) & Weight ($w_i$) \\
        \midrule
        Topic Classification (0.92) & Episode Completion (0.65) & 0.4 \\
        Response Time (0.88) & User Retention (0.78) & 0.3 \\
        Voice Recognition (0.95) & Content Relevance (0.70) & 0.3 \\
        \bottomrule
    \end{tabular}
\end{table}

\subsection{Pattern Recognition System}
Building on the GAI, we implement a probabilistic framework to identify meaningful patterns in the relationship between task execution and goal achievement. This system models the complex dependencies between voice commands, content delivery, and user satisfaction:

\begin{equation}
    P(g|t) = \int P(g|h,t)P(h|t)\,dh, \quad g,h,t \in \mathcal{S}
    \label{eq:pattern}
\end{equation}

where:
- $g$ represents goal achievement states (e.g., sustained engagement)
- $t$ represents task execution states (e.g., command processing)
- $h$ represents hidden intermediate states (e.g., user satisfaction)
- $\mathcal{S}$ is the state space of the system

The pattern recognition process operates in three stages:
1) Task-State Mapping: $P(h|t)$ models how task execution influences system state
2) State-Goal Relationship: $P(g|h,t)$ captures how states contribute to goals
3) Integration: The integral combines these relationships across all possible states

For example, in audiobook navigation:
- $t$: Voice command for chapter navigation
- $h$: User attention and comprehension levels
- $g$: Complete book consumption with good retention
\subsection{Implementation Methodology}
The implementation of CAT follows a structured four-phase approach, as depicted in Figure~\ref{fig:3_implementation}. Each phase builds upon insights from the previous stages, creating a continuous improvement cycle tailored to voice-activated audio services.

\subsubsection{Phase I: Strategic Decomposition}
We systematically decompose high-level strategic objectives into measurable components through a formal decomposition function:

\begin{equation}
    S_i = f(G_i, M_i, C_i) = w \circ m \circ d(G_i, M_i, C_i)
    \label{eq:decomposition}
\end{equation}

where:
- $S_i$ represents the strategic objective $i$
- $G_i$ denotes associated goal metrics
- $M_i$ captures measurement criteria
- $C_i$ represents contextual constraints
- $d$: decomposition operator
- $m$: metric mapping function
- $w$: weighting function

For instance, in audio content discovery:
\begin{itemize}
    \item $S_i$: "Enhance user satisfaction in podcast discovery"
    \item $G_i$: \{engagement duration, content completion rate, return frequency\}
    \item $M_i$: \{session length > 30min, episode completion > 80\%, weekly active days > 4\}
    \item $C_i$: \{network bandwidth, device capabilities, content licensing restrictions\}
\end{itemize}

The decomposition process:
1. $d(G_i, M_i, C_i)$ breaks down "user satisfaction" into specific metrics
2. $m(\cdot)$ maps these metrics to measurable criteria
3. $w(\cdot)$ assigns weights based on their importance to the strategic objective

\subsubsection{Phase II: Baseline Assessment}
We establish initial performance benchmarks through the assessment function:

\begin{equation}
    B(s) = \frac{\sum(v_i \cdot p_i)}{\sum v_i}, \quad p_i \in [0,1], v_i > 0, \sum v_i = 1
    \label{eq:baseline}
\end{equation}

where:
- $s$ represents the system state
- $p_i$ are normalized performance metrics
- $v_i$ are importance weights

\begin{table}[h]
    \centering
    \caption{Baseline Assessment Example}
    \begin{tabular}{lll}
        \toprule
        Metric ($p_i$) & Weight ($v_i$) & Description \\
        \midrule
        0.85 & 0.4 & Daily active users \\
        0.72 & 0.3 & Avg. session duration \\
        0.68 & 0.3 & New artist discovery rate \\
        \bottomrule
    \end{tabular}
\end{table}

\subsubsection{Phase III: Dynamic Evaluation}
We continuously assess system performance using the Goal Achievement Index (GAI) from Equation \ref{eq:gai} and the pattern recognition system from Equation \ref{eq:pattern}. This phase enables real-time identification of misalignments between task execution and goal achievement.

\subsubsection{Phase IV: Adaptive Optimization}
We frame the optimization process as a Markov Decision Process (MDP):

\begin{equation}
    \pi^*(s,g) = \argmax_{a \in A} \mathbb{E}\left[\sum_{t=0}^{\infty} \gamma^t U(s_t,a_t,g)\right]
    \label{eq:mdp}
\end{equation}

where:
- $\pi^*$ is the optimal policy
- $s$ is the current state
- $g$ is the goal state
- $A$ is the action space
- $\gamma \in [0,1)$ is the discount factor
- $U$ is the utility function

In the context of audiobook services, this could represent optimizing chapter recommendations based on user engagement patterns.

\subsection{Framework Integration}
The CAT framework integrates these components through an overarching function:

\begin{equation}
    I(c,f) = \lambda_1Q(c,f) + \lambda_2C(c,f) + \lambda_3P(c,f), \quad \sum \lambda_i = 1
    \label{eq:integration}
\end{equation}

where:
- $c$ is the current system configuration
- $f$ is the target framework implementation
- $Q(c,f)$ measures output quality (e.g., GAI improvement)
- $C(c,f)$ quantifies computational cost
- $P(c,f)$ evaluates system performance (e.g., response time)
- $\lambda_i$ are non-negative weights

This integration function allows for balanced optimization across multiple objectives. For instance, in a voice-activated music service:
- $Q$: Improvement in music discovery satisfaction
- $C$: Additional computational resources required
- $P$: Impact on voice command response time

Through this integrated approach, the CAT framework provides a comprehensive methodology for evaluating and optimizing Agentic AI systems in voice-activated audio services. It addresses the complex challenges of goal-oriented assessment while maintaining the flexibility to adapt to evolving user needs and technological capabilities.

\section{Results and Analysis}
This section presents a comprehensive analysis of the CAT framework's performance when applied to synthetic Alexa+ audio services data, demonstrating its potential in bridging the gap between task execution and goal achievement in Agentic AI systems.

\subsection{Experimental Design and Setup}
To evaluate the CAT framework's capabilities, we designed a large-scale experimental study using synthetic data modeling voice-activated audio experiences. This approach allowed us to explore a wide range of interaction scenarios while maintaining privacy and experimental control.

\subsubsection{Synthetic Data Generation}
Our synthetic dataset was generated using a probabilistic model based on aggregate statistics from real-world voice-activated audio services. The model incorporates:
\begin{itemize}
    \item User behavior patterns (e.g., listening times, content preferences)
    \item Voice interaction characteristics (e.g., command types, error rates)
    \item Content engagement metrics (e.g., completion rates, exploration patterns)
\end{itemize}

We acknowledge that this synthetic data has limitations and may not capture all nuances of real-world interactions. However, it allows us to test the framework's capabilities across a broader range of scenarios than would be possible with production data.

\subsubsection{Experimental Setup}
Our validation encompassed 15 synthetic audio service configurations, each representing a unique combination of:
\begin{itemize}
    \item User demographics (age groups, listening habits)
    \item Content libraries (genre distributions, content length variations)
    \item Interaction modalities (voice-only, voice+screen, etc.)
\end{itemize}

These configurations were designed to cover a spectrum of potential Alexa+ deployments across music streaming, podcast discovery, and audiobook consumption domains. The experiment simulated a six-month period, following a randomized controlled trial methodology. Systems were assigned to either the CAT framework or a control group (standard Alexa+ without CAT enhancements).

The synthetic dataset comprised over 1 million generated interaction points, providing a foundation for evaluating the framework's performance across diverse operational contexts.

\subsection{Framework Performance Analysis}

\subsubsection{Goal-Task Alignment Performance}
To quantify the GTAM's effectiveness, we employ the Goal Achievement Ratio (GAR):

\begin{equation}
\text{GAR} = \frac{\text{IQR}(l_i - \hat{l}_i)}{\text{IQR}(l_i)}
\end{equation}

Where $l_i$ represents actual listening engagement, $\hat{l}_i$ denotes predicted listening engagement, and IQR is the interquartile range. We chose IQR over standard deviation to better handle potential skewness in the engagement metrics.

Our analysis shows significant improvements in voice-audio engagement achieved by the CAT framework compared to the baseline Alexa+ system. Table~\ref{tab:1_performance_metrics} details the quantitative improvements across all measured dimensions.

\subsubsection{Pattern Recognition Effectiveness}
We evaluate pattern recognition performance using a modified F1 score:

\begin{equation}
F1^* = \frac{2 \cdot \text{precision} \cdot \text{recall}}{\text{precision} + \text{recall}} \cdot (1 + H(p))
\end{equation}

Where $H(p)$ represents the entropy of the pattern distribution, providing a measure of pattern complexity that goes beyond simple content type counting.

\subsection{Domain-Specific Results}
In the music streaming domain, as shown in Table~\ref{tab:1_performance_metrics}, content discovery rates increased from 2.8\% to 6.9\%, while daily listening time improved from 85 to 187 minutes. These improvements suggest that the CAT framework effectively addresses the challenge of aligning task execution with user goals in music discovery scenarios.

The podcast discovery system demonstrated similar improvements, with episode completion rates increasing from 35\% to 82\% and new show exploration rising from 2.3\% to 5.8\%. These metrics were derived from our synthetic interaction data, which modeled user behavior patterns based on typical podcast consumption characteristics.

For audiobook services, the framework achieved improvements in completion rates (28\% to 65\%) and genre exploration (1.9\% to 4.7\%). The synthetic data for this domain incorporated specific audiobook consumption patterns, including chapter-level engagement metrics, session duration distributions, and genre transition matrices.

\subsection{Cross-Domain Applicability}
The framework's ability to transfer knowledge across audio domains was evaluated using our transfer learning metric:

\begin{equation}
\tau(s_1, s_2) = \cos(v(s_1), v(s_2)) \cdot \left(1 - \frac{|\Psi(s_1) - \Psi(s_2)|}{\max(\Psi(s_1), \Psi(s_2))}\right)
\end{equation}

The feature vectors $v(s_1)$ and $v(s_2)$ were constructed using domain-specific components such as duration preferences, genre affinities, and content exploration rates for music; episode length tolerance and topic interests for podcasts; and completion tendency and narrator affinities for audiobooks.

Table~\ref{tab:2_cross_domain} presents the transfer success rates observed, which reflect the inherent similarities and differences between domains in our synthetic model. The highest transfer success was observed from music to podcast services (87\%), likely due to shared characteristics in content discovery patterns. The transfer from podcast to audiobook services showed moderate success (73\%), leveraging similarities in long-form content consumption. The lower success rate in audio to music transfer (68\%) indicates challenges in adapting between long-form and short-form content consumption patterns.

These transfer rates were calculated using a hold-out validation set of synthetic interactions, with success defined as achieving within 15\% of the source domain's performance metrics in the target domain. While the synthetic nature of our data allows for systematic testing across different scenarios, we acknowledge that real-world transfer performance may vary. To address this limitation, our synthetic data generation process incorporated varying levels of noise and domain-specific characteristics to approximate real-world challenges in cross-domain adaptation.

\begin{table*}[htbp]
\centering
\caption{CAT Framework Performance Metrics in Synthetic Alexa+ Audio Services}
\label{tab:1_performance_metrics}
\begin{tabularx}{\textwidth}{@{}>{\raggedright\arraybackslash}XcccX@{}}
\toprule
\makecell{Do- \\ main} & Metrics & \makecell{Base \\ line} & \makecell{With \\ CAT} & \makecell{Im- \\ prov.} \\
\midrule
\makecell{Music \\ Streaming}   & Daily Listening Time  & 85 min & 187 min & +120\% \\
                                & Content Discovery Rate & 2.8\%  & 6.9\%   & +146\% \\
                                & Service Retention      & 52\%   & 89\%    & +71\%  \\
\midrule
\makecell{Podcast \\ Discovery} & Episode Completion     & 35\%   & 82\%    & +134\% \\
                                & New Show Exploration   & 2.3\%  & 5.8\%   & +152\% \\
                                & Monthly Active Users   & 45\%   & 78\%    & +73\%  \\
\midrule
\makecell{Audiobook \\ Services} & Completion Rate       & 28\%   & 65\%    & +132\% \\
                                & Genre Exploration      & 1.9\%  & 4.7\%   & +147\% \\
                                & User Retention         & 41\%   & 76\%    & +85\%  \\
\bottomrule
\end{tabularx}
\end{table*}

\subsection{Statistical Validation}
To validate these improvements, we calculated Cohen's $d$ effect size:

\begin{equation}
    d = \frac{\mu_\text{CAT} - \mu_\text{baseline}}{\sigma_\text{pooled}}
\end{equation}

Note that we specifically compare against the baseline Alexa+ system without CAT enhancements to isolate the framework's impact. Table~\ref{tab:3_statistical_validation} presents the statistical analysis across key metrics, with all improvements showing statistical significance ($p < 0.001$).

\begin{table*}[htbp]
    \centering
    \caption{Statistical Validation of CAT Framework Performance (Synthetic Data)}
    \label{tab:3_statistical_validation}
    \begin{tabularx}{\textwidth}{@{}l>{\raggedright\arraybackslash}Xccc@{}}
        \toprule
        Metric & Effect Size (d) & p-value & Sample Size & Confidence Interval \\
        \midrule
        Daily Listening Time & 1.84 & < 0.001 & 250,000 & [1.79, 1.89] \\
        Content Discovery & 2.13 & < 0.001 & 250,000 & [2.08, 2.18] \\
        Service Retention & 1.62 & < 0.001 & 250,000 & [1.57, 1.67] \\
        Episode Completion & 1.95 & < 0.001 & 150,000 & [1.89, 2.01] \\
        Genre Exploration & 1.78 & < 0.001 & 100,000 & [1.72, 1.84] \\
        \bottomrule
    \end{tabularx}
\end{table*}

\begin{table}[h]
    \centering
    \caption{Feature Vector Components by Domain (Synthetic Data Model)}
    \label{tab:feature_components}
    \begin{tabularx}{\columnwidth}{@{}lX@{}}
        \toprule
        Domain & Features \\
        \midrule
        Music & Duration preferences, genre affinities, tempo preferences, artist exploration rates \\
        Podcast & Episode length tolerance, topic interests, host preferences, update frequency engagement \\
        Audiobook & Completion tendency, genre preferences, narrator affinities, listening speed settings \\
        \bottomrule
    \end{tabularx}
\end{table}

The transfer success rates observed (Table~\ref{tab:2_cross_domain}) reflect the inherent similarities and differences between domains in our synthetic model.

\begin{table*}[htbp]
\centering
\caption{Cross-Domain Transfer Success with Synthetic Data Parameters}
\label{tab:2_cross_domain}
\begin{tabularx}{\textwidth}{@{}lcc>{\centering\arraybackslash}c>{\raggedright\arraybackslash}X@{}}
\toprule
Transfer Type & \makecell{Success Rate \\ (95\% CI)} & \makecell{Sample \\ Size} & \makecell{Data Gen \\ Param} & Key Features \\
\midrule
Music → Podcast & 87±3.2\% & 250K & \makecell{\{$\sigma$=0.1, \\ $\alpha$=0.8\}} & Genre-topic mapping, duration patterns \\
Podcast → Audio & 73±2.8\% & 250K & \makecell{\{$\sigma$=0.15, \\ $\alpha$=0.7\}} & Content length, engagement cycles \\
Audio → Music & 68±3.1\% & 250K & \makecell{\{$\sigma$=0.2, \\ $\alpha$=0.6\}} & Attention spans, session boundaries \\
\bottomrule
\multicolumn{5}{@{}l@{}}{\small $\sigma$: noise parameter, $\alpha$: temporal correlation factor}
\end{tabularx}
\end{table*}

These transfer rates were calculated using a hold-out validation set of synthetic interactions, with success defined as achieving within 15\% of the source domain's performance metrics in the target domain.

The synthetic nature of our data allows us to systematically test these transfers across different scenarios, though we acknowledge that real-world transfer performance may vary. To address this limitation, our synthetic data generation process incorporated varying levels of noise and domain-specific characteristics to approximate real-world challenges in cross-domain adaptation.

\begin{table*}[htbp]
\centering
\caption{CAT Framework Performance Metrics in Synthetic Alexa+ Audio Services with Confidence Bounds}
\label{tab:1b_performance_metrics}
\begin{tabularx}{\textwidth}{@{}p{2.5cm}>{\raggedright\arraybackslash}Xcc>{\centering\arraybackslash}X@{}}
\toprule
Domain & Metrics & \makecell{Baseline \\ (95\% CI)} & \makecell{With CAT \\ (95\% CI)} & \makecell{Improvement} \\
\midrule
\makecell{Music \\ Streaming} 
& Daily Listening Time    & 85±4.2 min & 187±6.3 min & +120\% \\
& Content Discovery Rate  & 2.8±0.3\%  & 6.9±0.4\%   & +146\% \\
& Service Retention       & 52±2.1\%   & 89±2.8\%    & +71\%  \\
\midrule
\makecell{Podcast \\ Discovery}
& Episode Completion      & 35±1.8\%   & 82±2.4\%    & +134\% \\
& New Show Exploration    & 2.3±0.2\%  & 5.8±0.3\%   & +152\% \\
& Monthly Active Users    & 45±2.0\%   & 78±2.5\%    & +73\%  \\
\midrule
\makecell{Audiobook \\ Services}
& Completion Rate         & 28±1.5\%   & 65±2.1\%    & +132\% \\
& Genre Exploration       & 1.9±0.2\%  & 4.7±0.3\%   & +147\% \\
& User Retention          & 41±1.9\%   & 76±2.3\%    & +85\%  \\
\bottomrule
\multicolumn{5}{@{}l@{}}{\small Note: Results based on synthetic data with 1M interaction points. CI = Confidence Interval}
\end{tabularx}
\end{table*}

\section{Discussion}
Our experimental validation of the CAT framework using synthetic Alexa+ audio service data demonstrates the potential for improving alignment between task execution and user goals in voice-activated AI systems. The framework's architecture, illustrated in Figure~\ref{fig:1_framework_goal}, provides a foundation for systematic evaluation of goal-oriented AI systems.

\subsection{Core Framework Performance}
The Goal-Task Alignment Mechanism, central to our framework's architecture, shows promise in enhancing audio content discovery and user engagement in our synthetic experiments. The pattern analysis system, depicted in Figure~\ref{fig:2_pattern_evolution}, analyzes multiple dimensions of user interaction simultaneously, including temporal patterns in content consumption, cross-domain engagement behaviors, and contextual preferences. This multi-dimensional analysis enables more nuanced evaluation of system performance beyond traditional metrics.

The quantitative results from our synthetic data evaluation, detailed in Table~\ref{tab:1b_performance_metrics}, indicate improvements across all tested audio services. In particular, the content discovery and user engagement metrics suggest the potential for significant advances over traditional evaluation approaches. These improvements were consistent across different synthetic user profiles and interaction patterns.

\subsection{Implementation Considerations}
Our implementation approach, structured according to Figure~\ref{fig:3_implementation}, demonstrates consistency across different audio service contexts in our synthetic evaluation environment. The evaluation framework successfully handled diverse content types and user interaction patterns, suggesting potential applicability across various audio content delivery scenarios.

\subsection{Cross-Domain Learning and Adaptation}
Analysis of our synthetic data reveals interesting patterns in cross-domain knowledge transfer. The framework showed varying degrees of effectiveness in transferring learning between different audio domains, with particularly strong performance in scenarios where content consumption patterns share similar characteristics. This suggests the potential for leveraging insights across different types of audio experiences, though real-world validation would be necessary to confirm these findings.

\subsection{Methodological Insights}
Through our synthetic data experiments, we identified several key factors that influence the effectiveness of goal-oriented evaluation:

1. The importance of contextual understanding in voice interactions
2. The role of temporal patterns in content engagement
3. The impact of cross-domain relationships on user experience
4. The significance of balanced metric weighting in goal assessment

\subsection{Limitations and Future Work}
Our synthetic evaluation revealed several areas requiring additional research attention:

1. Complex Query Understanding: Our experiments indicate challenges in evaluating systems handling multi-intent voice queries
2. Content Pattern Recognition: More sophisticated approaches may be needed for identifying subtle patterns in content preferences
3. Cross-Domain Adaptation: Further investigation of transfer learning mechanisms across audio domains
4. Evaluation Metrics: Development of more comprehensive metrics for assessing goal achievement

\subsection{Summary of Findings}
The experimental results from our synthetic data evaluation suggest that the CAT framework offers a promising approach to assessing goal-oriented AI systems. While these results are encouraging, we acknowledge that real-world application would require additional validation and refinement of the framework's components.
\section{Conclusion and Future Work}

The Creative Adversarial Testing (CAT) framework represents a novel approach to evaluating voice-activated audio experiences, shifting focus from command recognition accuracy to user engagement and content discovery effectiveness. Our comprehensive study using synthetic data modeled after Alexa+ music streaming, podcast discovery, and audiobook consumption scenarios demonstrates the framework's potential to bridge the gap between voice interaction performance and intended audio content engagement outcomes.

The key contributions of the CAT framework, as evaluated in our synthetic Alexa+ audio services environment, can be summarized as follows:

1. The quantification and optimization of the relationship between voice command accuracy and user engagement through the Goal-Task Alignment Mechanism (GTAM), as evidenced by the improvements shown in Table~\ref{tab:1b_performance_metrics}

2. The implementation of a multi-dimensional listening pattern recognition system that outperformed traditional static recommendation methods in our synthetic tests, as conceptualized in Figure~\ref{fig:2_pattern_evolution}

3. The exploration of cross-service applicability, with promising results in knowledge transfer across music streaming, podcast discovery, and audiobook consumption scenarios in our synthetic environment

4. The development of a scalable, modular architecture that demonstrates potential for integration across different audio content types, as illustrated in Figure~\ref{fig:3_implementation}

These insights from our synthetic data experiments underpin the CAT framework's potential. The GTAM mechanism shows promise in quantifying and optimizing the relationship between task performance and goal achievement, while the adaptive pattern recognition system demonstrates advantages over static evaluation approaches in our tests. Furthermore, the framework showed consistent performance across diverse synthetic audio application scenarios.

Building on these initial findings, we have identified several critical areas for future research:

1. Voice Personalization: Enhancing the framework's ability to adapt to individual user preferences and interaction styles
2. Advanced Content Discovery: Developing more sophisticated algorithms for identifying and recommending relevant content across audio domains
3. Context Awareness: Improving the system's understanding of user context to provide more relevant responses and recommendations
4. Cross-Domain Transfer Learning: Further exploring the potential for knowledge transfer between different audio content types

To advance this work, we propose several focus areas for the research community:

- Standardization: Developing unified protocols for goal definition and evaluation in audio content engagement
- Benchmark Datasets: Creating comprehensive, privacy-preserving datasets for goal-oriented testing in voice-activated services
- Integration Tools: Developing open-source tools for goal-task alignment analysis in AI frameworks

As we look ahead, several challenges remain in the development and evaluation of goal-oriented frameworks for voice-activated audio services. These range from resolving intent ambiguity in complex voice queries to enabling seamless navigation of long-form content. Each challenge presents a distinct research opportunity, with potential solutions including context-aware natural language processing and adaptive learning models.

In summary, while our work with the CAT framework is based on synthetic data and requires real-world validation, it suggests potential for fundamentally changing how we evaluate and improve AI systems in the audio domain. By focusing on goal achievement rather than task performance alone, such approaches could better align voice-activated technologies with intended user outcomes. The framework's performance across diverse synthetic audio service scenarios suggests broad applicability and potential impact on user engagement in voice-activated AI systems.

Realizing this vision will require collaborative efforts from both research and industry communities in the field of voice-activated AI and audio content delivery. By embracing the challenge of goal-oriented AI system evaluation in audio services, we can work towards intelligent systems that more effectively serve the audio content needs and preferences of users.

\section{Acknowledgments}
To Atif Khan, for providing me the opportunity to explore and contribute to this work. And to Jonathan Foley, for his thoughtful proofreading and helpful suggestions.

\bibliographystyle{plainnat}   
\bibliography{arxiv}

\begin{thebibliography}{11}
\providecommand{\natexlab}[1]{#1}
\providecommand{\url}[1]{\texttt{#1}}
\expandafter\ifx\csname urlstyle\endcsname\relax
  \providecommand{\doi}[1]{doi: #1}\else
  \providecommand{\doi}{doi: \begingroup \urlstyle{rm}\Url}\fi

\bibitem[Deng et~al.(2023)Deng, Gu, Zheng, Chen, Stevens, Wang, Sun, and Su]{deng2023mind2web}
Xiang Deng, Yu~Gu, Boyuan Zheng, Shijie Chen, Samuel Stevens, Boshi Wang, Huan Sun, and Yu~Su.
\newblock Mind2web: Towards a generalist agent for the web.
\newblock In \emph{Advances in Neural Information Processing Systems 36}, pages 2565--2580, 2023.
\newblock Datasets and Benchmarks Track.

\bibitem[Kim et~al.(2024)Kim, Jang, Logeswaran, Kim, Kim, Lee, and Lee]{kim2024prospector}
Byoungjip Kim, Youngsoo Jang, Lajanugen Logeswaran, Geon-Hyeong Kim, Yu~Jin Kim, Honglak Lee, and Moontae Lee.
\newblock Prospector: Improving llm agents with self-asking and trajectory ranking.
\newblock In \emph{Findings of the Association for Computational Linguistics: EMNLP 2024}, pages 14958--14976, 2024.

\bibitem[Kim et~al.(2023)Kim, Sohn, Logeswaran, Shim, and Lee]{kim2023multiprompter}
Dong-Ki Kim, Sungryull Sohn, Lajanugen Logeswaran, Dongsub Shim, and Honglak Lee.
\newblock Multiprompter: Cooperative prompt optimization with multi-agent reinforcement learning.
\newblock In \emph{Third Workshop on Efficient Natural Language and Speech Processing (ENLSP), NeurIPS 2023}, 2023.

\bibitem[Lin et~al.(2023)Lin, Fu, Yang, Brahman, Huang, Bhagavatula, Ammanabrolu, Choi, and Ren]{lin2023swiftsage}
Bill~Yuchen Lin, Yicheng Fu, Karina Yang, Faezeh Brahman, Shiyu Huang, Chandra Bhagavatula, Prithviraj Ammanabrolu, Yejin Choi, and Xiang Ren.
\newblock Swiftsage: A generative agent with fast and slow thinking for complex interactive tasks.
\newblock In \emph{Advances in Neural Information Processing Systems 36}, pages 10754--10766, 2023.

\bibitem[Shinn et~al.(2023)Shinn, Cassano, Gopinath, Narasimhan, and Yao]{shinn2023reflexion}
Noah Shinn, Federico Cassano, Ashwin Gopinath, Karthik Narasimhan, and Shunyu Yao.
\newblock Reflexion: Language agents with verbal reinforcement learning.
\newblock In \emph{Advances in Neural Information Processing Systems 36}, pages 18421--18433, 2023.

\bibitem[Sun et~al.(2023)Sun, Zhuang, Kong, Dai, and Zhang]{sun2023adaplanner}
Haotian Sun, Yuchen Zhuang, Lingkai Kong, Bo~Dai, and Chao Zhang.
\newblock Adaplanner: Adaptive planning from feedback with language models.
\newblock In \emph{Advances in Neural Information Processing Systems 36}, pages 6253--6266, 2023.

\bibitem[Wang et~al.(2025)Wang, Ma, Feng, Zhang, Yang, Zhang, Chen, Tang, Chen, Lin, Zhao, Wei, and Wen]{wang2025survey}
Lei Wang, Chen Ma, Xueyang Feng, Zeyu Zhang, Hao Yang, Jingsen Zhang, Zhi-Yuan Chen, Jiakai Tang, Xu~Chen, Yankai Lin, Wayne~Xin Zhao, Zhewei Wei, and Ji-Rong Wen.
\newblock A survey on large language model based autonomous agents.
\newblock \emph{Frontiers of Computer Science}, 0:\penalty0 1--42, 2025.
\newblock \doi{10.1007/s11704-024-40231-1}.

\bibitem[Wang et~al.(2023)Wang, Cai, Chen, Liu, Ma, Liang, and CraftJarvis]{wang2023deps}
Zihao Wang, Shaofei Cai, Guanzhou Chen, Anji Liu, Xiaojian Ma, Yitao Liang, and Team CraftJarvis.
\newblock Describe, explain, plan and select: Interactive planning with large language models enables open-world multi-task agents.
\newblock In \emph{Advances in Neural Information Processing Systems 36}, pages 19254--19266, 2023.

\bibitem[Wei et~al.(2022)Wei, Wang, Schuurmans, Bosma, Ichter, Xia, Chi, Le, and Zhou]{wei2022cot}
Jason Wei, Xuezhi Wang, Dale Schuurmans, Maarten Bosma, Brian Ichter, Fei Xia, Ed~Chi, Quoc~V. Le, and Denny Zhou.
\newblock Chain-of-thought prompting elicits reasoning in large language models.
\newblock In \emph{Advances in Neural Information Processing Systems 35}, pages 24868--24884, 2022.

\bibitem[Yang et~al.(2023)Yang, Prabhakar, Narasimhan, and Yao]{yang2023intercode}
John Yang, Akshara Prabhakar, Karthik Narasimhan, and Shunyu Yao.
\newblock Intercode: Standardizing and benchmarking interactive coding with execution feedback.
\newblock In \emph{Advances in Neural Information Processing Systems 36}, pages 2283--2295, 2023.
\newblock Datasets and Benchmarks Track.

\bibitem[Yao et~al.(2023)Yao, Yu, Zhao, Shafran, Griffiths, Cao, and Narasimhan]{yao2023tree}
Shunyu Yao, Dian Yu, Jeffrey Zhao, Izhak Shafran, Thomas~L. Griffiths, Yuan Cao, and Karthik Narasimhan.
\newblock Tree of thoughts: Deliberate problem solving with large language models.
\newblock In \emph{Advances in Neural Information Processing Systems 36}, pages 2365--2379, 2023.

\end{thebibliography}

\end{document}